\RequirePackage{fix-cm}
\documentclass[twocolumn]{svjour3}          
\smartqed  

\usepackage{scrtime}

\usepackage{enumitem}
\usepackage{setspace}
\usepackage{wrapfig}
\usepackage{colortbl}
\usepackage{float}
\usepackage{graphicx}
\usepackage{amsmath}
\usepackage{amssymb}
\usepackage{times}
\usepackage{epsfig}
\usepackage[utf8]{inputenc}
\usepackage{multirow}
\usepackage{algorithm}
\usepackage[normalem]{ulem}
\usepackage{adjustbox}
\usepackage{cuted}
\usepackage{flushend}

\usepackage[dont-mess-around]{fnpct}
\usepackage{xcolor}

\usepackage{etoolbox}
\usepackage{hyperref}
\newcommand*{\affaddr}[1]{#1} 
\newcommand*{\affmark}[1][*]{\textsuperscript{#1}}

\journalname{Neural Computing and Applications}
\begin{document}

\title{Cross-Region Building Counting in Satellite Imagery using Counting Consistency}

\author{%
Muaaz Zakria \protect\affmark[1] \and Hamza Rawal\affmark[1] \and Waqas Sultani\affmark[1] \and Mohsen Ali \affmark[1]
}
\authorrunning{Muaaz Zakria \and Hamza Rawal\and Waqas Sultani \and Mohsen Ali}

\institute{
            \\
              \email{waqas.sultani@itu.edu.pk}           
          \and
            \at
              \\
\affaddr{\affmark[1]Intelligent Machines Lab, Information Technology University, Lahore}
}

\maketitle
\begin{abstract} Estimating the number of buildings in any geographical region is a vital component of urban analysis, disaster management, and public policy decision. Deep learning methods for building localization and counting in satellite imagery, can serve as a viable and cheap alternative. However, these algorithms suffer performance degradation when applied to the regions on which they have not been trained.  Current large datasets mostly cover the developed regions and collecting such datasets for every region is a costly, time-consuming,  and difficult endeavor.  In this paper,  we propose an unsupervised domain adaptation method for counting buildings where we use a labeled source domain (developed regions) and adapt the trained model on an unlabeled target domain (developing regions). We initially align distribution maps across domains by aligning the output space distribution through adversarial loss. We then exploit counting consistency constraints, within-image count consistency, and across-image count consistency, to decrease the domain shift. Within-image consistency enforces that the building count in the whole image should be greater than or equal to the count in any of its sub-image.  Across-image consistency constraint enforces that if an image contains considerably more buildings than the other image, then their sub-images shall also have the same order. These two constraints encourage the behavior to be consistent across and within the images, regardless of the scale.  To evaluate the performance of our proposed approach, we collected and annotated a large-scale dataset consisting of challenging South Asian regions having higher building densities and irregular structures as compared to existing datasets. We perform extensive experiments to verify the efficacy of our approach and report improvements of approximately 7\% to 20\% over the competitive baseline methods.
The dataset and code are available here: { \href{https://github.com/intelligentMachines-ITU/domain-Adaptive-Building-Counting}{https://github.com/intelligentMachines-ITU/ domain-Adaptive-Building-Counting}.}
\end{abstract}

\section{Introduction}
 
The precise and accurate estimation of the number of buildings is vital for many tasks, such as monitoring economic well-being \cite{yeh2020using}, planning aid for a natural or a man-made disaster-stricken region \cite{ali2020destruction}, analyzing poverty \cite{jean2016combining}, and predicting the vitality of a city \cite{scepanovic2021jane}. Over the years, there has been an effort to approximate the population size by estimating the number of buildings in an area through satellite imagery and through land use/cover data \cite{wang2018mapping}. Building counting acts as an indicator of important metrics such as population density \cite{qiu2003modeling,harvey2002estimating}, and power usage \cite{li2005using}. This information is generally collected through various censuses and surveys or their fusion, requiring costly, expansive, and time-consuming efforts. One way to tackle this challenge is to make the process cost-effective and labor-saving by using deep learning-based methods on satellite images to get an automatic estimate of building counts \cite{shakeel2019deep}   {or through the extraction of buildings \cite{xia2019geosay} }.

The problem of counting objects in an image has been studied extensively in the last few years. 
    The ubiquitous nature of the counting problem is exhibited in the variety of research works that {deal with counting cells in the petri dish \cite{marsden2018people}, estimating crowd size  \cite{liu2019context}, and counting the buildings  \cite{shakeel2019deep}. In the literature, counting has been performed mainly through segmentation \cite{li2008estimating}, clustering \cite{rabaud2006counting}, and regression-based methods \cite{liu2019context,zhang2016single,sam2017switching,tu2008unified,onoro2016towards}.  As compared to counting in normal images, counting the number of buildings (or in short, building counting) from satellite imagery is less explored \cite{saeedi2008automatic}}. 

\begin{figure*}[t]
\includegraphics[width = \linewidth]{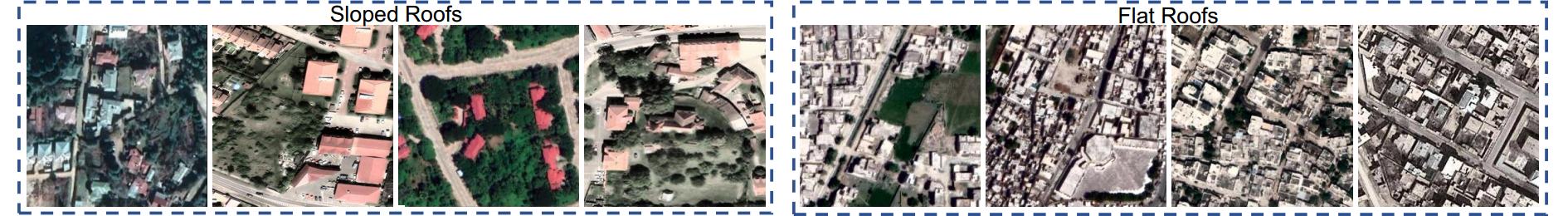}
\caption{The left block shows the images captured from colder regions and have sloped roofs while the images in the right block show flat roofs as those areas belong to hotter regions.}
\label{fig: coldhot}
\end{figure*}

 The training data on which a deep learning model is trained is called the source domain (data) and while the one on which it is to be tested is called the target domain (data). Ideally, source and target domain data distributions should be the same, however, in real-world applications,  source, and target domain distributions are different. This is more apparent in the case of the building counting problem, since depending upon the region, climate, and culture, the building structures are different from each other.
In Figure \ref{fig: coldhot}, for example, we show images from the colder region having buildings with sloped roofs, whereas the other ones have flat roofs. Similarly density of the neighborhood, building material, time of image acquisition,  quality, and resolution of the aerial/satellite imagery, can affect the robustness of the building count. The gap between the distributions of two (source and target) domains, called domain shift, is the reason for this failure.  It is well known that the deep learning-based methods, even when trained on large datasets fail to generalize to the new domain \cite{iqbal2020mlsl,benjdira2019unsupervised,hossain2020domain,iqbal2020weakly,subhani2020learning}.

To overcome the limitation due to domain shift, domain adaptation strategies have been applied to numerous problems including crowd counting  \cite{hossain2020domain,benjdira2019unsupervised,subhani2020learning,iqbal2020mlsl,iqbal2020weakly}. 
Domain adaptation in building counting is a relatively unexplored problem and more focus of these domain adaptation problems is on semantic segmentation or object detection etc \cite{iqbal2020weakly}. 

Recently various remote sensing datasets have been introduced for deep learning applications \cite{xia2018dota,van2018spacenet}. However, these datasets majorly cover regions from the developed world. When a deep learning model trained on developed regions, is tested on developing or under-developed regions, it declines in performance due to domain shift. 

In this paper, we propose to tackle this performance decline in building counting across regions (from developing to under-developed regions). In our case, the source dataset contains regions from developed countries and the target dataset contains developing regions.  \textit{Since we do not assume the availability of ground truth data of the target training dataset, we call our approach an unsupervised domain adaptation method}. Given the satellite image, we first automatically obtain building density maps employing \cite{liu2019context}. 
The total building count in an image is produced by summing the whole density map. 
Learning to predict the map forces the model to learn to localize the buildings it is trying to count. However, due to the domain shift, the density maps predicted on the target domain lack structure. To address this, we propose to use adversarial learning which forces the model to learn to produce density maps that are indistinguishable across the source and target domains.  
Furthermore, we design a problem-specific strategy to align two domains for improved building counting. Specifically, we propose two counting consistency constraints, within-image count consistency, and across-image count consistency, to help decrease the domain shift in an unsupervised way.
Both of these consistencies should naturally occur in any area. 
Within-image consistency encompasses the logic that the number of buildings in the whole image should be greater than or equal to the number of buildings in any of the sub-region captured in the image.
Across-image consistency constraint on the other hand enforces that if a region contains considerably more number of buildings than other regions, then a considerably large sub-region of the former one will also have more number of buildings than the sub-region of the latter one. 
These two constraints encourage the behavior to be consistent across and within the images, regardless of the scale.

To summarize, the following are the contributions of our work:
\begin{itemize}
\item We attempt to address a new problem of cross-region building counting and localization.
 
\item We propose two problem-specific constraints, count consistency, to direct the unsupervised domain adaptation process. These constraints i.e., the within-image and across-image count consistency constraints force the model to learn a generalized representation of buildings. 
\item We collect a new large-scale and challenging dataset for building localization and counting with a focus on South Asian regions having irregular building structures.
    \item We perform extensive experiments to show the effectiveness of our proposed approach and the efficacy of the collected dataset.

\end{itemize}

We will cover the related works in section \ref{sec:related}, explore the used datasets in section \ref{sec:datasets}, explain our adopted methodologies in section \ref{sec:methodology}, detail out our implementation and discuss our results in section \ref{sec:implementation}, and finally conclude in section \ref{sec:conclusion}.


\section{Related Work} \label{sec:related}
 {In our proposed approach, we tackle the problem of cross-region building counting and localization using counting consistency constraints based on ranking loss and introduce a new dataset. Hence, below we discuss the works related to object counting, ranking, domain adaptation, and remote sensing datasets.}\newline
\noindent\textbf{Object Counting}: Counting objects of interest through computer vision has been done in several application areas which include counting people, animals, fruits, buildings, etc. 
Counting crowd from images was performed in the early days using detection based methods \cite{wu2005detection,xiong2017spatiotemporal,li2008estimating}.
Clustering-based methods have also been employed to count people in crowded scenes \cite{rabaud2006counting,tu2008unified}. However, regression-based counting methods \cite{liu2019context,onoro2016towards,sam2017switching} have generally produced more state-of-the-art results than the other mentioned methods.
Recently, Liu et al. \cite{liu2019context} have predicted crowd density by encoding the contextual information contained within various scales. To account for the fact that the appropriate scale varies over the image, Kang et al. \cite{kang2018crowd} proposed to weight the generated density maps differently at different scales.   { Whereas, Li et al. \cite{li2020effective} performed crowd counting using multi-resolution context and image quality assessment-guided training.}
To count building from satellite imagery, Shakeel et al., \cite{shakeel2019deep} proposed a regression model using attention-based re-weighting.

 {Detecting and counting fruits is of unmatched importance in agriculture \cite{rahnemoonfar2017deep,liu2018robust,zabawa2020counting}. Rahnemoonfar et al. \cite{rahnemoonfar2017deep} and  Liu et al. \cite{liu2018robust} used deep learning-based methodologies to count fruits in images. Similarly, Zabawa et al. \cite{zabawa2020counting} used convolutional neural networks to perform semantic segmentation to detect grapevine berries in images and then count them using a connected component algorithm. 
The problem of counting vehicles has been tackled using various methodologies in literature \cite{guerrero2015extremely,onoro2016towards,zhang2017fcn}. }
 
\noindent\textbf{Ranking}: 
The ranking makes sure that the ordering of a list of items is in the correct order. The framework of ranking has been applied to solve various problems of computer vision including improving the counting of crowds in congested scenes using a ranking loss \cite{liu2018leveraging}, anomaly detection in surveillance videos where the ranking loss was used to localize the anomalies during the training \cite{sultani2018real} and detection of features by employing a deep ranking framework \cite{wang2014learning}. 

\noindent\textbf{Domain Adaptation}:
Domain adaptation is the process of minimizing the effects of a domain shift that arises when training and testing data is drawn from different distributions. Domain adaptation has been used to solve several problems such as  {object classification \cite{zhang2020domain},   {detection \cite{zheng2020cross,soviany2021curriculum,su2021multi} }, semantic segmentation \cite{tasar2020colormapgan,iqbal2020mlsl,benjdira2019unsupervised,iqbal2020weakly}, {person re-identification \cite{hou2021unsupervised} } and crowd counting \cite{hossain2020domain,li2019coda}} .  {Domain adaptation has been addressed by \cite{mozafari2017cluster} by proposing a new latent sub-domain discovery model for dividing the target domain into sub-domains by considering them a cluster while bridging the domain gap}. A weakly supervised domain adaptation network for latent space and output space has been proposed by \cite{iqbal2020weakly} to diminish the cross-domain gap in satellite and aerial imagery for performing semantic segmentation of built-up areas. Hossain et al., \cite{hossain2020domain} used domain adaptation for crowd counting where they used semi-supervised domain adaptation using a limited number of labeled images from target data. Finally, they have minimized maximum mean discrepancy (MMD) loss between the generated density maps of source and target. 
In addition to semi or weakly supervised domain adaptation, some recent works also address unsupervised domain adaptation  { \cite{liang2018aggregating,liang2019exploring,hou2021unsupervised,zhou2021cluster} }.    Moreover, domain adaptation in remote sensing image classification was also addressed by Zhang et al. \cite{zhang2020domain} and Liu et al. \cite{liu2020multikernel} using unsupervised transfer learning.
 {In addition to above cited works, domain shift problems have also been addressed in recommendation systems by \cite{li2015context} using context-aware bandits, by \cite{li2016collaborative} using Collaborative Filtering Bandits, and by similar bandits \cite{korda2016distributed,mahadik2020fast}.}

\noindent\textbf{Datasets for Remote Sensing}:
Several datasets have been introduced for remote sensing applications such as object detection or built-areas detection. One of the most popular datasets for object detection is the xView dataset \cite{lam2018xview}. 
 It contains bounding box annotations of objects of multifarious classes. The xView dataset consists of a total of 1 million object instances that come under 60 classes. It covers a land area of 1415$km^2$. A building detection dataset was released by the SpaceNet \cite{van2018spacenet}, covering the areas of Rio De Janeiro, Las Vegas, Paris, Shanghai, and Khartoum. A semantic labeling benchmark dataset was launched by ISPRS \cite{isprs} which contained 2D semantic labels in high quality of two cities in Germany. For the problem of counting built structures from satellite imagery, a large and diverse dataset was introduced by \cite{shakeel2019deep} covering urban, hilly, and desert regions. 
Another important dataset was proposed by \cite{ali2020destruction} to detect destructed sites due to natural and man-made disasters from satellite imagery. Similarly, for object detection in aerial imagery, a large-scale dataset, DOTA, was put forward by \cite{xia2018dota}. 

In contrast to the above-mentioned works, our approach focuses on counting buildings in satellite imaging across different regions. Instead of employing within-image counts ranking for self-training \cite{liu2018leveraging}, we have used across and within-image counts ranking for unsupervised domain adaptation across different regions. Furthermore, we have introduced a new dataset and annotations to demonstrate the efficacy of the proposed approach.

\begin{figure*}[t]
\centering
\includegraphics[width=0.9\textwidth]{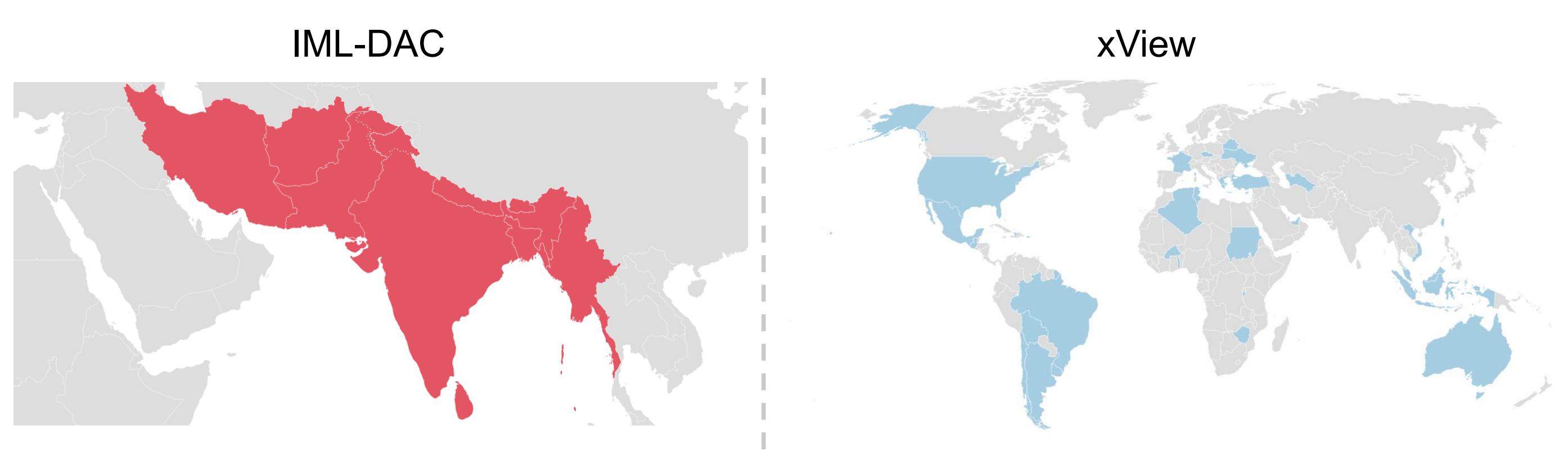}
\caption{Side-by-side comparison of geographical locations of IML-DAC and xView datasets.}
\label{fig:xviewarea}
\end{figure*}

\begin{figure*}
\centering
\includegraphics[width=0.95\textwidth]{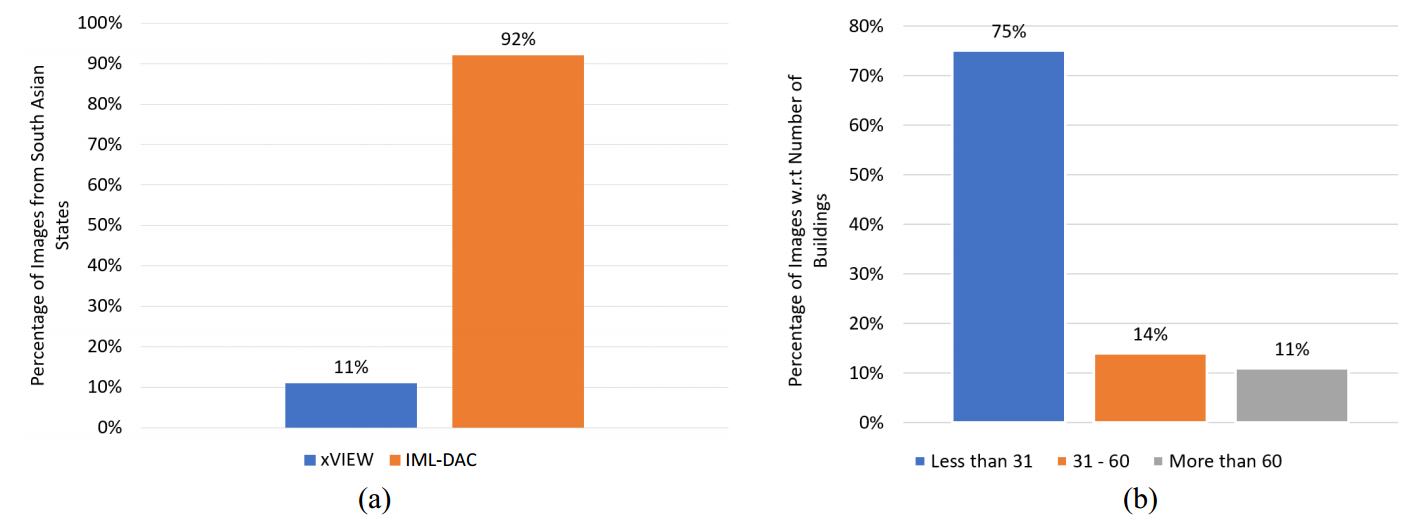}
\caption{(a) compares the percentage of South Asian regions contained in xView and IML-DAC. Our dataset contains 81\% more images from South Asian regions than xView dataset. (b) shows the distribution of images of xView and IML-DAC datasets with respect to the number of buildings contained in them.}
\label{fig:comparisonxm}
\end{figure*}
\begin{figure*}
\centering
\includegraphics[width=1.0\textwidth]{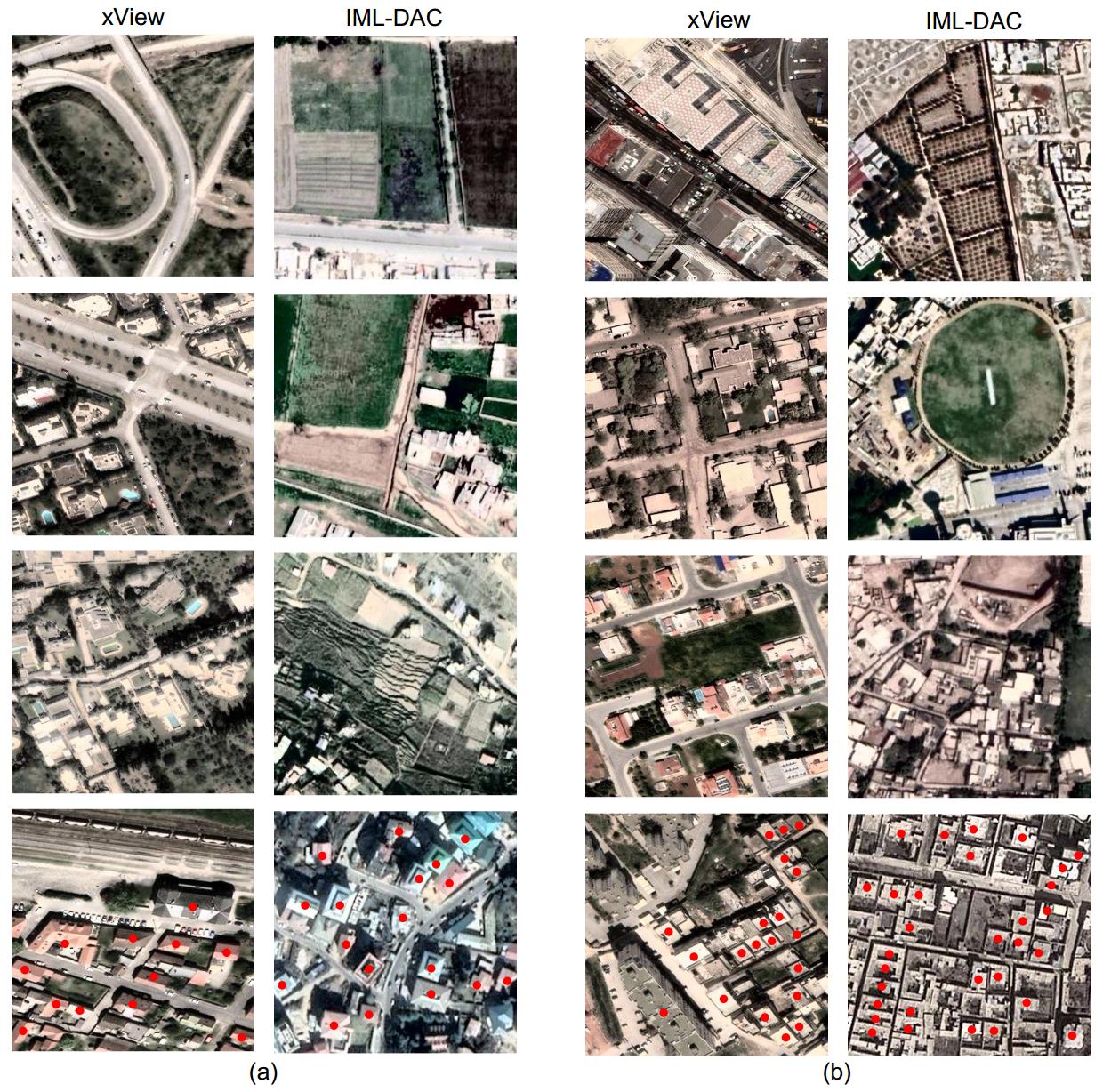}
\caption{Comparison of images from both datasets. (a) shows the images of  xView and IML-DAC for similar building counts. Similarly (b) highlights the difference in building structures. In the last row, we show the annotated points on the buildings. The images are histogram equalized.}
\label{fig: comparison}
\end{figure*}
\noindent\section{Dataset Preparation} \label{sec:datasets}

\begin{figure}[t]
 \centering
\includegraphics[width=0.4\textwidth]{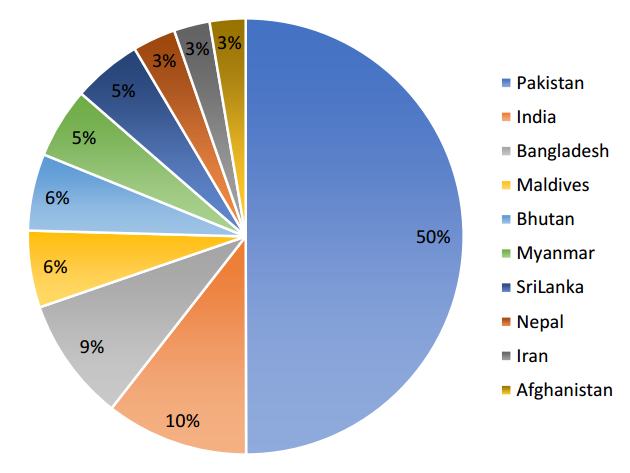}
\caption{Distribution of IML-DAC dataset. The majority of images have been collected from Pakistan across its various cities.}
\label{fig:areawise}
\end{figure}

\noindent\textbf{xView}: 
We appropriated xView \cite{lam2018xview} for the building counting task utilizing it as a \textit{source} dataset. 
The motivation behind using this dataset is that it is distributed across various parts of the world to detect objects ranging across 60 classes which also includes buildings\footnote{We have not used  DOTA \cite{xia2018dota} since it does not contain buildings class.}. The xView dataset originally contained 847 training images along with ground truth. The ground truth is available in geoJSON format. The fields in it contained the bounding box label ID, a unique ID for image strips, image filename, coordinates of bounding boxes, and longitude-latitude information of bounding boxes.
To generate our desired annotations, we first selected the bounding boxes which covered \textbf{Buildings} only. 
Afterward, we use original ground truth coordinates to generate the points which are in the center of these buildings.  { After that, we generate non-overlapping patches of size 500 $\times$ 500 pixels and selected a total of 4935 patches}.
The division of these patches with respect to the number of buildings contained in them is shown in Figure \ref{fig:comparisonxm}(b). Note the xView is geographically biased towards developed regions/countries. There are a few images from developing countries that we did not use during training the model.
\newline\newline
\noindent\textbf{IML-DAC}: To evaluate the accuracy of the proposed approach in cross-region building counting, in our experiments, we use the xView as a source (train) dataset and IML-DAC and South Asian regions of xView as target (test) datasets. The geographical locations of areas from which xView and IML-DAC are collected are presented in Figure \ref{fig:xviewarea}. Figure \ref{fig:comparisonxm}(a) compares the percentage of South Asian regions contained in both datasets and Figure \ref{fig:comparisonxm}(b) shows the distribution of images of xView and IML-DAC with respect to the number of buildings contained in them. Figure \ref{fig:areawise} communicates a better understanding of the distribution of our dataset with respect to the percentage of images from each region. 

In Figure \ref{fig: comparison}, we demonstrate a  comparison of both datasets with respect to their counts and structures. Figure \ref{fig: comparison}(a) shows the images (side by side) of xView and IML-DAC having similar building counts. It can be observed that buildings in xView are well-placed and distant while there is no proper planning for building placements in the IML-DAC dataset. Figure \ref{fig: comparison}(b) highlights the difference in structures of both datasets. Most images in xView contain tall buildings while in IML-DAC, the majority of buildings are small in size or either built of non-concrete material. Since our main goal is to count buildings, the images in  IML-DAC are annotated through a dotted annotation on each building. Some typical examples of annotations are shown in the bottom row of  Figure \ref{fig: comparison}. Note that as mentioned in the `xView' section, dotted annotations for xView are extracted from the bounding box annotations provided by the original authors of xView. 

\section{Methodology} 
\label{sec:methodology}
In the following section, we provide the details of each component used in our pipelines and explain all the design choices behind them. 

\subsection{Preliminaries}
Let us define the source and target datasets to be $\mathcal{D}^s=\{(I_i^s, l_i^s), i = 1 \dots N^s\}$ and $\mathcal{D}^t=\{(I_j^t), j=1 \dots N^t\}$. 
Where  $I_i^s$ and $I_j^t$ are the satellite imagery patches from the source and target datasets, and  {$N^s$ and  $N^t$ are the total number of source and target data image patches respectively}. Note that in this paper, we use satellite images from developed regions as a \textit{source} dataset and satellite images from developing regions as \textit{target} dataset. For each image patch $I_i^s$ in the source domain dataset, we have a ground-truth list $l_i^s$ of locations where the buildings are present.
Using \cite{liu2019context}  the ground-truth density map $D_i^s$ is created for each source sample, such that a Gaussian is centered on each location in  $l_i^s$ and the variance of the Gaussian depends upon how far away the other buildings are from the current one. 

\begin{figure}[t]
\includegraphics[width=0.50\textwidth]{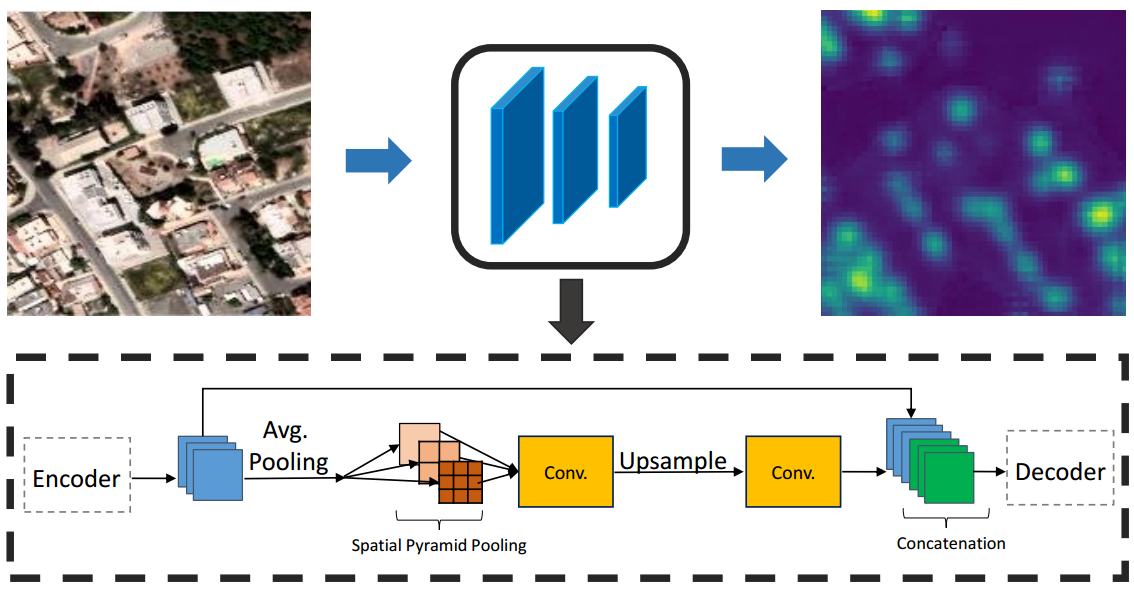}
\caption{Generating density maps of buildings from satellite images using   Context-Aware  Convolutional  Network  (CACN) \cite{liu2019context}.}
\label{fig:CACN}
\end{figure}

Since the proposed approach works on the building density maps, we use the Context-Aware Convolutional Network (CACN) \cite{liu2019context} based on counting pipeline $f_c$. 
Originally this network was introduced to count people in images of crowded scenes. 
The motivation behind using CACN for counting buildings is the fact that it can encode the contextual information contained within multiple scales in an adaptive manner by incorporating spatial pyramid pooling.
Such spatial and contextual variations are also visible in the building dataset. CACN is trained over the source domain using the ground-truth density maps  $D_i^s$'s .
\begin{equation}
\mathcal{L}_{MSE}^s = \frac{1}{2N^s} \sum_{i=1}^{N^s} \left \| D_i^s - f_c(I_{i}^s) \right \|_{2}^{2},
\label{Eq:MSE}
\end{equation}
where $\mathcal{L}_{MSE}^s$ represents the mean square loss on the source dataset, $N^s$ is the batch size chosen for training,  $D_i^s$ is the ground truth density map and $f_c(I_{i}^s)$ is the predicted density map of image patch $I_{i}^s$.  Figure \ref{fig:CACN} shows the example of a building density map generated by CACN. 
During inference, the predicted count is the summation of the density map. This source-only model serves as a baseline upon which we add our proposed modules to conduct progressive performance enhancement.

\subsection{Unsupervised Domain Alignment}

The source-only trained model from the previous section performs poorly on the target domain. This decrease in performance is attributed to the domain gap between the source and target domains. We design an unsupervised domain adaptation strategy guided by adversarial feature alignment and the proposed consistency constraints.
The resultant model is robust to domain shift and is more generalized than the baseline source-only model.

\subsubsection{Distribution Map Alignment (DMA) using Adversarial Learning}

\begin{figure*}[t]
 \centering
\includegraphics[width=1.0\textwidth]{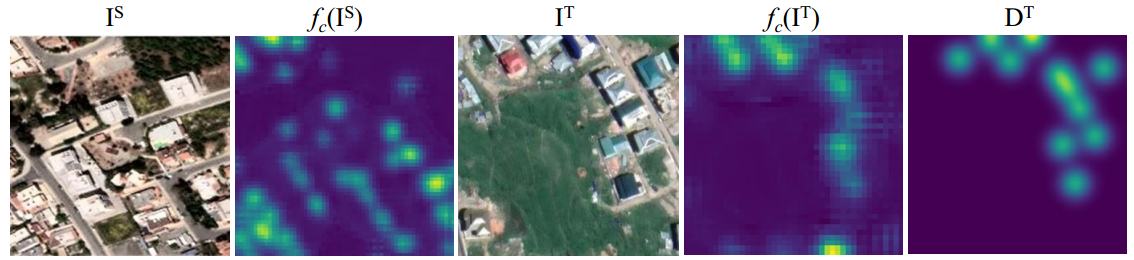}
\caption{The distribution map predicted for the source images by source trained model has a visible structure consisting of multiple Gaussians. However, prediction over the target image patch results in a distribution map that is more blurred and lacks structure. Here, $I^{S}$ is the source image patch, $f_c(I^{S})$ is the predicted distribution map of the source image patch from the source model, $I^{T}$ is the target image patch, $f_c(I^{T})$ is the predicted distribution map of the target image patch from the source model, $D^{T}$ is the ground-truth density map of the target image patch.}
\label{fig:distribution}
\end{figure*}
\begin{figure}[t]
\centering
\includegraphics[width=0.5\textwidth]{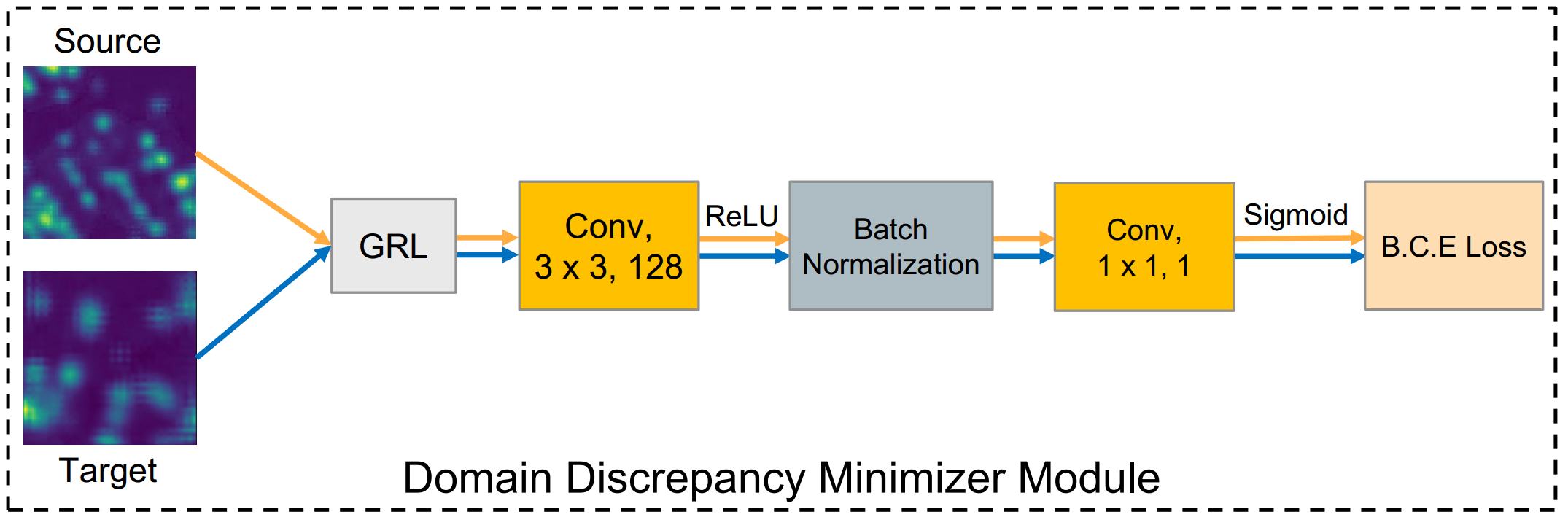}
\caption{Domain Discrepancy Minimizer Module: A discriminator and GRL are added after the network outputs for adversarial domain adaptation.}
\label{fig:ddmm}
\end{figure}
The counting pipeline, trained on the source dataset, produces the density map that has a structure similar to what is in the ground-truth maps. However, due to the domain shift, the distribution map produced for target data has visible artifacts as shown in Figure \ref{fig:distribution}.

\begin{figure}[t]
\centering
\includegraphics[width=0.5\textwidth]{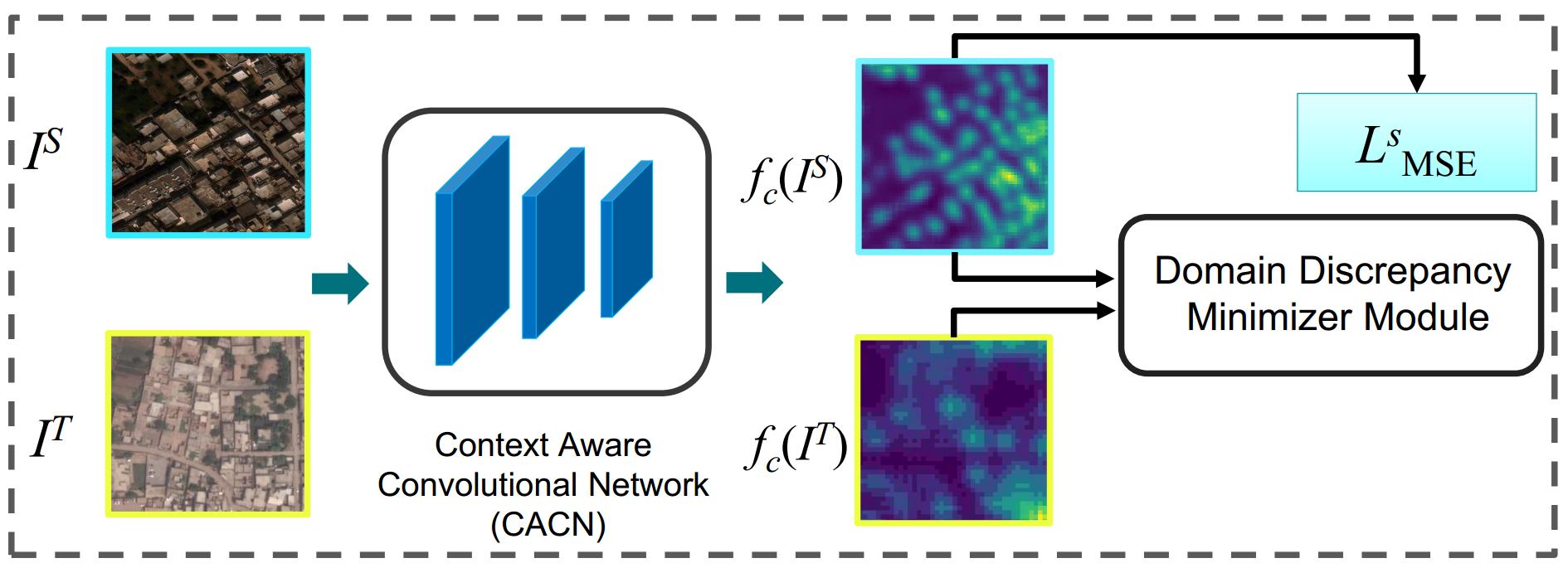}
\caption{This figure depicts distribution map alignment using adversarial learning. $I^{S}$ is the source image patch.  $I^{T}$ is the target image patch. $f_c()$ represents density maps of their respective images. $\mathcal{L}_{MSE}^s$ is the M.S.E loss between generated and ground truth density maps of the source image.
}
\label{fig:pipeline_1}
\end{figure}

The learned features of the network are biased by the supervised training of the source domain. Thus the network does not recognize the features that it needs to localize the center of the building in the target domain. 
To learn a better output space distribution for the target domain, we must enforce the network to learn to output similar distribution for the target as for the source domain. For this, we perform adversarial alignment over the output distribution map using a domain discrepancy minimizer module as depicted in Figure~\ref{fig:ddmm}. The process of adversarial alignment is described in Figure~\ref{fig:pipeline_1}.

The output distribution of a network for two domains is considered to be consistent if we cannot discriminate between the two distributions. A domain discriminator can be used to learn to classify whether the generated distribution map is for the source image or target image. The more discriminative the two outputs are, the less consistent the two distributions would be and vice versa. Thus we can see that there is an inverse relationship between the consistency of outputs of the network and the ability of the discriminator to distinguish them. In adversarial domain adaptation, outputs for the two domains are aligned by forcing the network that produces the outputs to generate similar distributions. Since there is a relation between the discriminator and the density-maps-generating network, we can use the former to adjust the latter by using the loss gradients of the discriminator. 

Ganin et al. \cite{ganin2015unsupervised} showed this the above-mentioned objective could be achieved by using a Gradient Reversal Layer (GRL) with a discriminator to learn domain invariant features. In this setting, the discriminator tries to distinguish the domains using a standard cross-entropy loss function. And as the training continues, we adjust the density-maps-generating network such that the discriminator is not able to distinguish the domains. This is where gradient reversal comes in. Gradients coming from the discriminator are reversed before being propagated to the density-maps-generating network. Thus, in effect, we are exploiting the reverse relation between the two where the gradients-updating-discriminator in one direction are updating the density-maps-generating-network in opposite direction. Thus while the discriminator is trying to distinguish the domains, the density-maps-generating network is now trying to generate outputs such that the discriminator is unable to distinguish the domains. As the training continues, the discriminator finds it more and more difficult to distinguish the domains which means that the density maps generated are more and more consistent. Thus, we can align the outputs of the network for the two domains.

In our case, the discriminator is a small network containing a couple of convolution layers and a batch norm layer as shown in Figure~\ref{fig:ddmm}. The discriminator minimizes the following standard classification loss function of Binary Cross Entropy:
\begin{equation}
\centering
\mathcal{L}_{B.C.E} = - \sum^{\textrm{N}_{c}}_{i=1} y_{i}\log(\hat{y}_{i}),
\label{eq:bce}
\end{equation}

where $\mathcal{L}_{B.C.E}$ is the Binary Cross Entropy Loss, $y$ is the label vector and $\textrm{N}_{c}$ is the number of classes which is two (source and target) in our case. The discriminator is applied after the output density maps and the GRL layer is used between the density maps outputs and the discriminator. It should also be noted that the discriminator and GRL are only added at training time to minimize the domain gap and are removed at test time. So, the total loss for the adversarial learning-based domain adaptation step is given as:
    \begin{equation}
    \mathcal{L}_{DMA} = \mathcal{L}_{MSE}^s  + \alpha \mathcal{L}_{B.C.E},
    \label{eq:dma}
    \end{equation} where $\mathcal{L}_{DMA}$ is distribution map alignment loss, $\alpha$ is the weighting factor and $\mathcal{L}_{MSE}^s$ is the mean square error computed over the source domain (Equation 1).

\subsubsection{Counting Consistency within Image}

 \begin{figure}[t]
\centering
\includegraphics[width=0.4\textwidth]{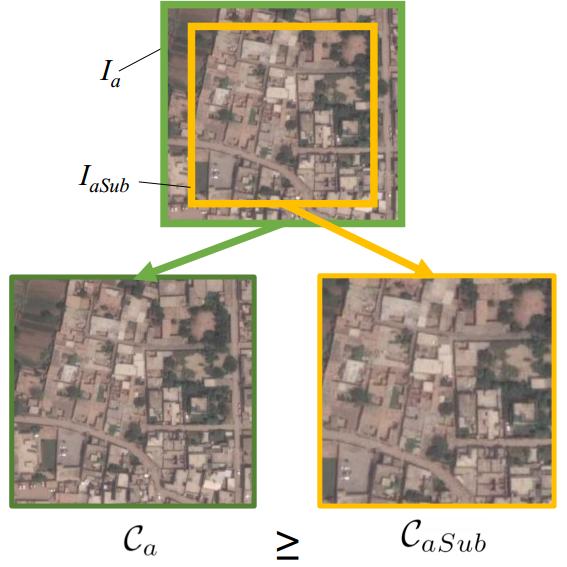}
\caption{Within-Image consistency constraint ensures that the count of $\mathcal{I}_a$ should be greater than or equal in value to the count of $\mathcal{I}_{aSub}$.}
\label{fig:WIC}
\end{figure}

\begin{figure}
\centering
\includegraphics[width=0.5\textwidth]{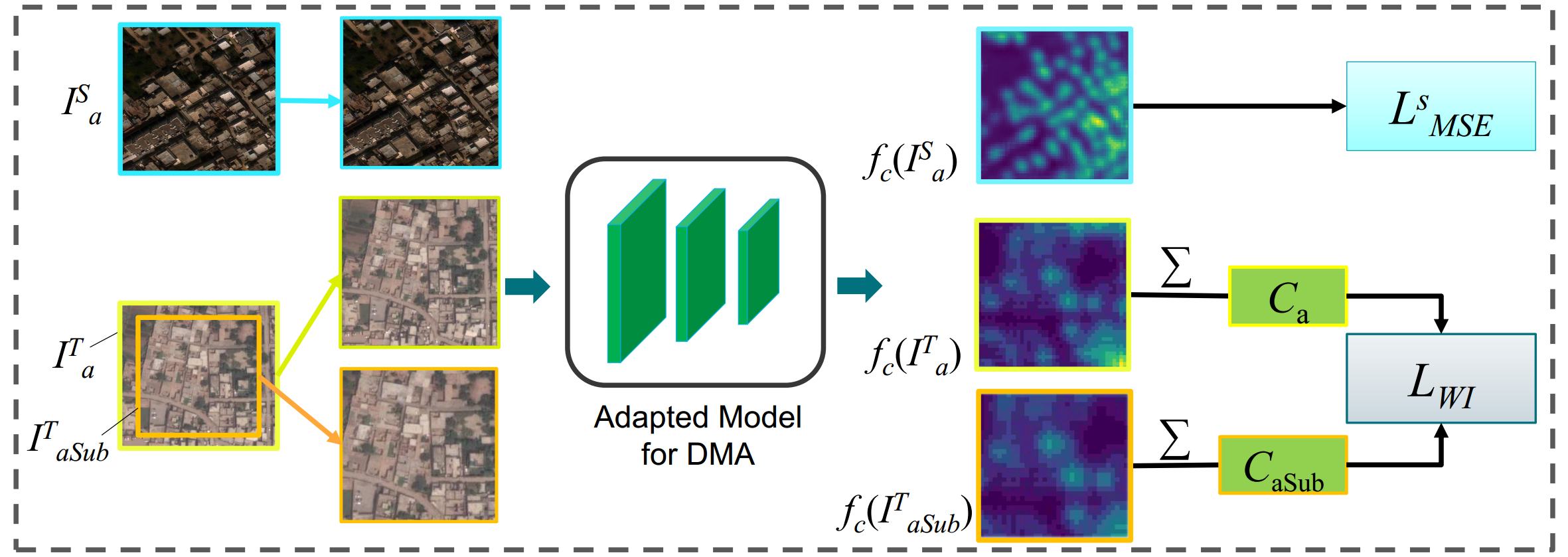}
\caption{This figure represents a learning framework employing image counting consistency constraint. $I^{S}_{a}$ is the source image.  $I^{T}_{a}$ is the target image. $I^{T}_{aSub}$ is the resized sub-image of $I^{T}_{a}$ . $f_c()$ represents density maps of their respective images. $\mathcal{L}_{MSE}^s$ is the M.S.E loss between generated and ground truth density map of the source image. $\mathcal{L}_{WI}$ is the within-image count consistency loss.
}
\label{fig:pipeline_2}
\end{figure}

To overcome the limitation of unlabeled target data, we design basic constraints that should be true for correct counting. In its basic form, any patch of the image cannot have more objects than the whole image.  To accomplish this, we have employed the ranking loss. Ranking loss has been used previously to constrain unsupervised deep learning-based methods \cite{liu2018leveraging}.
Let $\mathcal{C}_a=C(f_c(\mathcal{I}_a))$ be predicted number of buildings in the image patch $\mathcal{I}_a$ and the $I_{aSub}$ be the sub-patch (see Figure \ref{fig:WIC}) extracted from $\mathcal{I}_a$,  {and $\mathcal{C}_{aSub}=C(f_c(\mathcal{I}_{aSub}))$ is the predicted number of buildings in the sub-patch}.
The within-image count consistency loss is given as: \begin{equation}
\mathcal{L}_{WI} (\mathcal{I}_a, \mathcal{I}_{aSub}) = max(0, -(\mathcal{C}_a - \mathcal{C}_{aSub}) + m),
\label{eq:WIC}
\end{equation}
where $\mathcal{L}_{WI}$ is the within-image counting consistency loss, margin $m$  allows us to control the relaxation in the constraint.    
No loss is back-propagated in case of equal counts or when they are in the correct order as depicted in Equation  \ref{eq:WIC}. The total loss for this step is given as:
\begin{equation}
    \mathcal{L}_{CWI} = \mathcal{L}_{DMA} + \lambda_1 \mathcal{L}_{WI},
    \label{eq:L_WIC}
    \end{equation}where $\mathcal{L}_{CWI}$ is the total loss to compute within-image counting consistency, $\mathcal{L}_{DMA}$ is distribution map alignment loss, and $\lambda_1$ is the relative weight assigned. The whole process of learning employing counting consistency within images is shown in Figure~\ref{fig:pipeline_2}.

\subsubsection{Counting Consistency  Across Image}

\begin{figure}[t]
\centering
\includegraphics[width=0.5\textwidth]{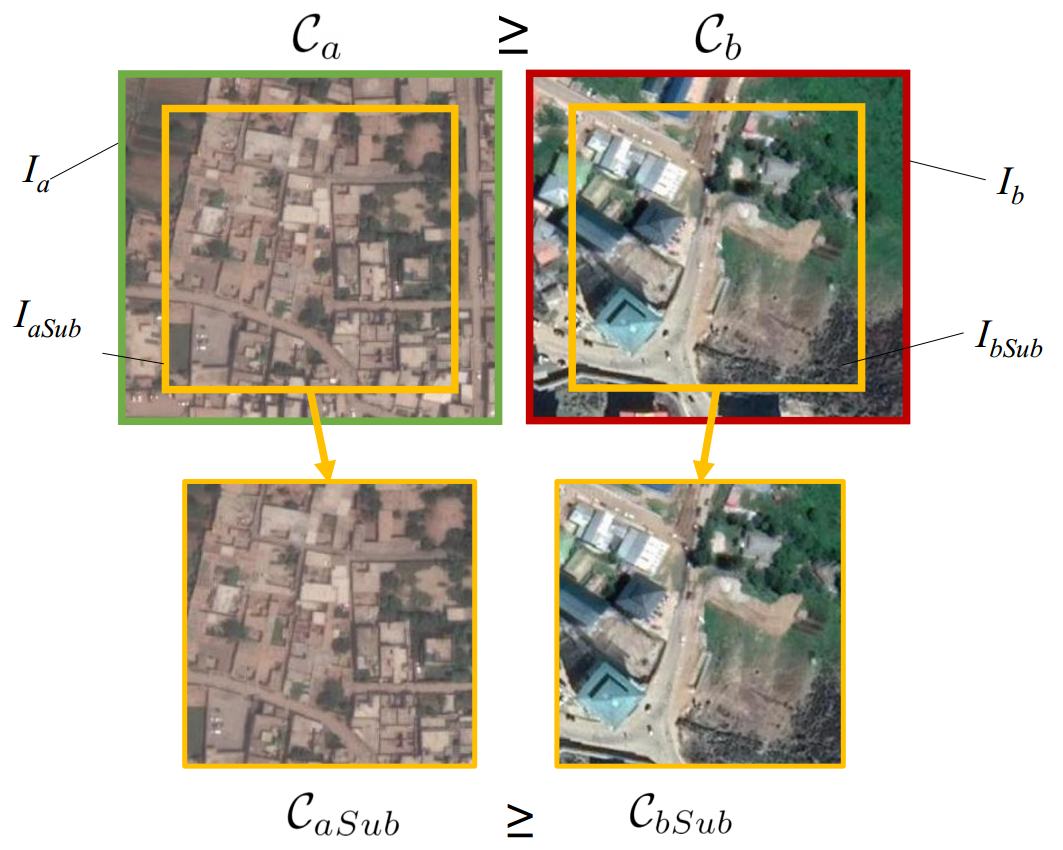}
\caption{Counting consistency across images ensures that if the predicted count of $\mathcal{I}_a$ is greater than or equal to the predicted count of $\mathcal{I}_b$, then the count of $\mathcal{I}_{aSub}$ should also be greater than or equal to the count of $\mathcal{I}_{bSub}$.}
\label{fig:across}
\end{figure}

Counting consistency loss within the same images is not a powerful enough constraint, as indicated by \cite{liu2018leveraging} who used it for   \textit{warmup} task before using the supervised learning to predict the counting. 
Therefore, a much stronger across-image-counting-consistency constraint is proposed to help in domain adaptation. 
The constraint states if image patch $\mathcal{I}_a$ has buildings substantially greater than the $\mathcal{I}_b$, any large-enough sub-patch of $\mathcal{I}_a$ will also contain buildings greater than or equal to the number of buildings in an equally large-enough sub-patch of $\mathcal{I}_b$ (see Figure \ref{fig:across}). 
Let $\mathcal{C}_a=C(f_c(\mathcal{I}_a))$ and $\mathcal{C}_b=C(f_c(\mathcal{I}_b))$ be predicted number of buildings in the patches $\mathcal{I}_a$ and $\mathcal{I}_b$, 
and $\mathcal{C}_{aSub}$ and $\mathcal{C}_{bSub}$ are the predicted number of buildings in the sub-patches respectively.

\begin{figure}[ht]
\centering
\includegraphics[width=0.5\textwidth]{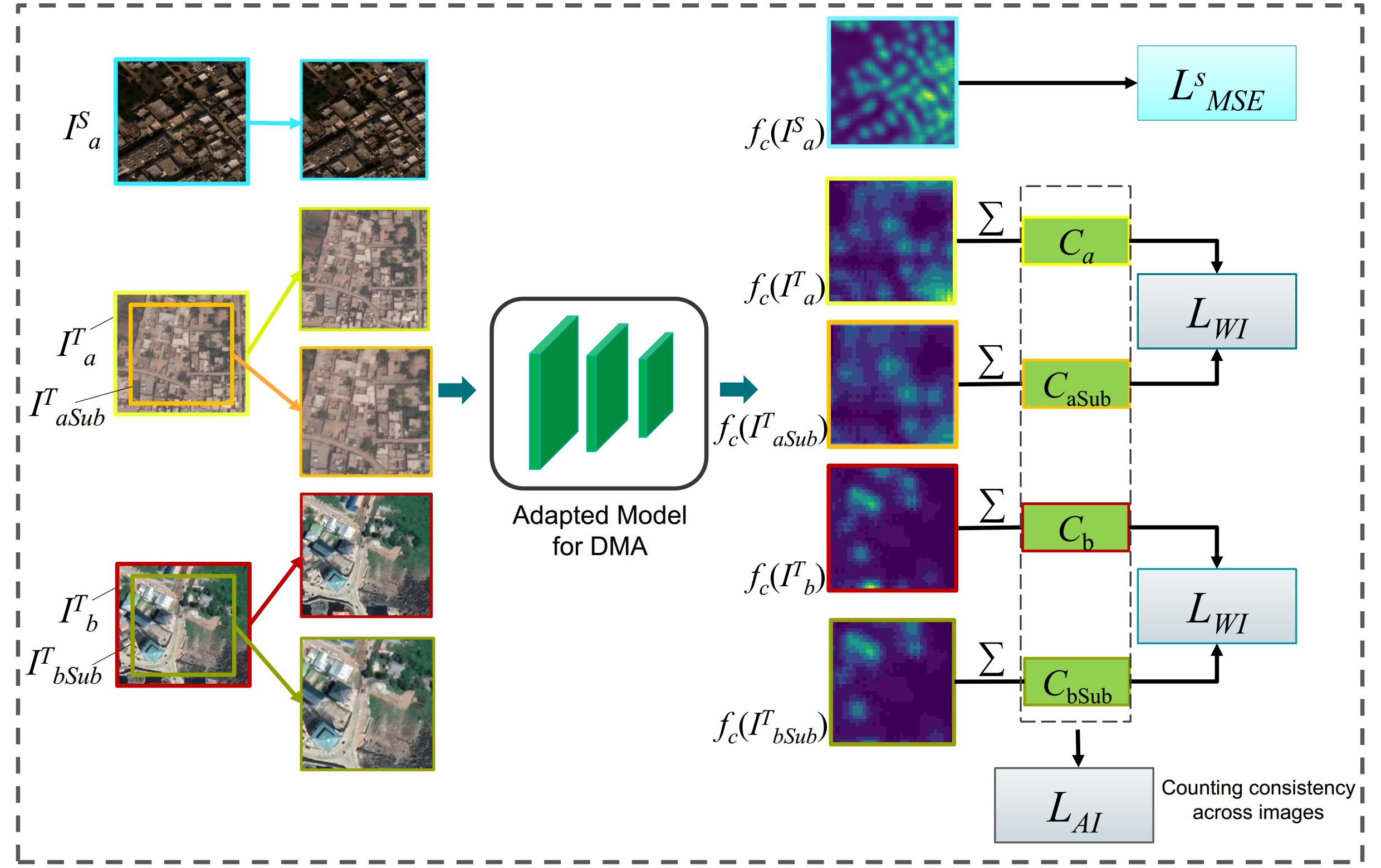}
\caption{This figure shows the learning framework using both within the image and across the image counting consistency constraints. $I^{S}_{a}$ is the source image patch.  $I^{T}_{a}$ , $I^{T}_{b}$ , $I^{T}_{aSub}$ , and $I^{T}_{bSub}$ are the target image patches and their respective resized sub-patches. $f_c()$ represents density maps of their respective patches. $\mathcal{L}_{MSE}^s$ is the M.S.E loss between generated and ground truth density map of the source image patch. $\mathcal{L}_{WI}$ is the within-image count consistency loss. $\mathcal{L}_{AI}$ is the across-image count consistency loss.
}
\label{fig:pipeline_3}
\end{figure} 

The constraint is represented as loss in the following equation,
\begin{eqnarray}
\small
 \mathcal{L}_{AI} (\mathcal{I}_a, \mathcal{I}_{b}) \ 
 \! = \!
\begin{cases}
max(0, -(\mathcal{C}_{aSub} - \mathcal{C}_{bSub}) + m)  & \!\!\! \mbox{if } \mathcal{C}_a \! \ge \! \mathcal{C}_b \\
max(0, -(\mathcal{C}_{bSub} - \mathcal{C}_{aSub}) + m) & \!\!\! \mbox{otherwise} 
\end{cases},
\label{eq:AIC}
\end{eqnarray}
where $\mathcal{L}_{AI}$ is the across-image counting consistency loss, $m$ is the margin. While implementing, $\mathcal{I}_a$ and $\mathcal{I}_b$ are chosen such that there is a larger than 5 difference in their count. To further keep this constraint true, the extracted sub-image is $80\%$ of the original image. The total loss in the case of the across-image-consistency loss is given as:
\begin{equation}
\begin{split}
\small
\mathcal{L}_{CAI} =\mathcal{L}_{DMA} + \lambda_1 \mathcal{L}_{WI}(\mathcal{I}_a, \mathcal{I}_{aSub}) 
\\+ \lambda_1 \mathcal{L}_{WI} (\mathcal{I}_b, \mathcal{I}_{bSub}) 
+ \lambda_2\mathcal{L}_{AI} (\mathcal{I}_a, \mathcal{I}_b),
\end{split}
\label{eq:L_AIC}
\end{equation}where $\mathcal{L}_{CAI}$ is the total loss to compute across-image counting consistency, $\mathcal{L}_{DMA}$ is distribution map alignment loss, $\mathcal{L}_{WI}$  computes within-image counting consistency loss, $\mathcal{L}_{AI}$  computes across-image counting consistency loss, $\lambda_1$ and $\lambda_2$ are the relative weights assigned.
The whole process of maintaining counting consistency across images is depicted in Figure~\ref{fig:pipeline_3}.

\section{Experiments} \label{sec:implementation}

\subsection{Implementation  details} 
Both source dataset and target datasets were divided into training, validation, and testing sets according to 60:20:20 ratios which made 2958 training images, 989 validation images, and 988 testing images.
The image patches of both source and target datasets have a size of $500 \times 500$ pixels each. 
As a pre-processing step, we performed histogram equalization on patches of both our source and target datasets such that they have the same contrast level.
We trained the source model on $2958$ training patches of xView dataset for $140$ epochs using Adam as an optimizer and by keeping a learning rate of $1e^{-4}$. 
$\mathcal{L}_{MSE}^s$ (Equation \ref{Eq:MSE}  )  is minimized between the predicted and ground truth density maps of the xView dataset. We kept the batch size to be equal to 26 patches. Validation was performed on $989$ patches of the validation set of xView. 

While adapting the source model for DMA, we have used the learning rate of 1e\textsuperscript{-5} and  $\alpha$ was set to 0.1. 
During this adaptation process, the training set of xView images and the unlabeled training set of IML-DAC were utilized. 
 In the experiments performed for within-image consistency only, the learning rate was kept at 1e\textsuperscript{-5} and the total number of epochs was set to be 50. 
 The relative weight $\lambda$ was chosen to be 45. During adaptation for both within-image consistency and across-the-images consistency, the learning rate and the number of epochs were 1e\textsuperscript{-5} and 50, while $\lambda_1$ and $\lambda_2$ were taken to be 45 and 1 respectively. The batch size was again kept as 26 for these two adaptation experiments. The number of training images from both our source and target datasets was kept the same to be 2958 respectively. Hence a total of 5916 training image patches were utilized in the adaptation processes. Testing of these adapted models was performed on 988 patches of the testing set of the IML-DAC dataset and on 230 patches of the South Asian subset of xView dataset.

\subsubsection{Computation Cost:}
 {Training is done on a single machine equipped with TitanX GPU having 12 GB memory. Source model training took approximately 12-13 hours to complete. 
Adaptation time for Distribution Map Alignment (DMA) took approximately 4 hours, within image Counting Consistency took approximately 4 hours, and the adaptation time for across-image counting consistency is approximately 6 hours.
Testing on a single image takes (on average) 4678 ms.}
\subsection{Evaluation Metrics:}
 To evaluate our approach, we compute Mean Relative Error (MRE) :
\begin{equation}
\begin{aligned} 
MRE = \frac{1}{N} \sum_{i=1}^{N}{(\frac{\left |\mathcal{C}_{GT}\left(I_i\right) - \mathcal{C}_{Pred}\left(I_i\right)  \right |  \times  100}{\mathcal{C}_{GT}\left(I_i\right)})}
\end{aligned}
\end{equation}
where $N$ is the number of image patches in our testing set, $C_{GT} \left(I_i\right)$ is the
ground truth count of buildings in the $i^{th}$ image and
$C_{Pred} \left(I_i\right)$ is the predicted count of buildings in the
$i$th image patch. MRE, also known as Mean Absolute Percentage Error (MAPE) \cite{de2016mean,tofallis2015better} is less susceptible to outliers in comparison to Mean Absolute Error (MAE) and the Root Mean Squared Error (RMSE)  \cite{de2016mean}.

\begin{table}[t]
\centering
\caption{This table demonstrates the hyper-parameters selection of our different experiments.}
\label{ablation}
\begin{adjustbox}{width=1.0\columnwidth,center}
\begin{tabular}{cccc} 
\hline
\textbf{Experiments} & \textbf{Parameters} & \textbf{MRE} & \begin{tabular}[c]{@{}c@{}}\textbf{$\omega$}\end{tabular} \\ 
\hline
\multirow{3}{*}{$\mathcal{L}_{DMA}$} & $\alpha$ = 0.05 & 32.91 & 118 \\ 
\cline{2-4}
 & $\alpha$ = 0.1 & \textbf{32.06} & \textbf{140} \\ 
\cline{2-4}
 & $\alpha$ = 0.15 & 32.65 & 106 \\ 
\hline
\multirow{4}{*}{$\mathcal{L}_{CWI}$} & $\alpha$ = 0.1, $\lambda_1$ = 35 & 31.25 & 890 \\ 
\cline{2-4}
 & $\alpha$ = 0.1, $\lambda_1$ = 40 & 31.98 & 815 \\ 
\cline{2-4}
 & $\alpha$ = 0.1, $\lambda_1$ = 45 & \textbf{30.16} & \textbf{1116} \\ 
\cline{2-4}
 & $\alpha$ = 0.1, $\lambda_1$ = 50 & 30.94 & 1077 \\ 
\hline
\multirow{4}{*}{$\mathcal{L}_{CAI}$} & $\alpha$ = 0.1, $\lambda_1$ = 45, $\lambda_2$ = 1/22 & 29.77 & 1098 \\ 
\cline{2-4}
 & $\alpha$ = 0.1, $\lambda_1$ = 45, $\lambda_2$ = 1/24 & 28.86 & 1246 \\ 
\cline{2-4}
 & $\alpha$ = 0.1, $\lambda_1$ = 45, $\lambda_2$ = 1/26 & \textbf{27.89} & \textbf{1464} \\ 
\cline{2-4}
 & $\alpha$ = 0.1, $\lambda_1$ = 45, $\lambda_2$ = 1/28 & 28.37 & 1321 \\
\hline
\end{tabular}

\end{adjustbox}
\end{table}

\subsection{Hyper-parameter Selection}

\begin{figure*}[t]
\centering
\includegraphics[width=0.8\textwidth]{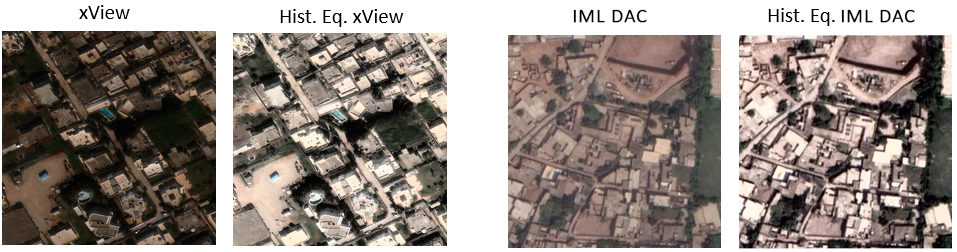}
\caption{Side-by-side comparison of patches, before and after histogram equalization, of IML-DAC and xView datasets.}
\label{fig:histcomparision}
\end{figure*}

Histogram equalization is utilized as a pre-processing step to improve the contrast level of the images and is applied to each image separately.
This preprocessing step improves the source trained model (that has not seen the target data) because it does not have to overcome the difference in contrast level in the two domains.
In Figure \ref{fig:histcomparision}, we have compared our patches of both xView and IML-DAC datasets, before and after applying histogram equalization.
Below we present the results before and after performing histogram equalization. As shown in Table \ref{tab:hist_result}, the source-only trained model performed 22.9\% better on the target dataset (IML-DAC) when histogram equalization was applied on patches.

 Optimal hyper-parameters are selected by training on a small part of the training dataset in the target domain (IML-DAC) and testing over the full training dataset. 
 {The hyper-parameters were selected according to $\omega$ which represents the number of image patches of the training set of our target data (IML-DAC) which followed within-image counting consistency and the MRE being computed on its testing set. Note that MRE is not used in choosing the parameters due to the assumption of the unavailability of ground truth density maps of the target training dataset. However, the correlation between the last two columns indicates the effectiveness of using with-in consistency loss for hyper-parameter selection.} 
 We start with $\mathcal{L}_{DMA}$ (Eq. \ref{eq:dma}) and iterate over different values of $\alpha$. 
 For each value of $\alpha$, the model is trained over a small dataset. The setting that results in the smallest within-image consistency loss over the full training dataset is chosen as the optimal value.   
Similarly, we find optimal value for $\lambda_1$, by minimizing $\mathcal{L}_{CWI}$ (Eq. \ref{eq:L_WIC}) and keeping $\alpha$ constant. 
For  $\lambda_2$ both the $\alpha$ and $\lambda_1$ are kept equal to optimal values picked in previous steps, as we minimize $\mathcal{L}_{CAI}$ (Eq. \ref{eq:L_AIC}). In Table \ref{ablation}, our different sets of experiments demonstrate the usefulness of our selected hyperparameters.

\subsection{Experimental Results}

\begin{figure*}[t]
\centering
\includegraphics[width=0.85\textwidth]{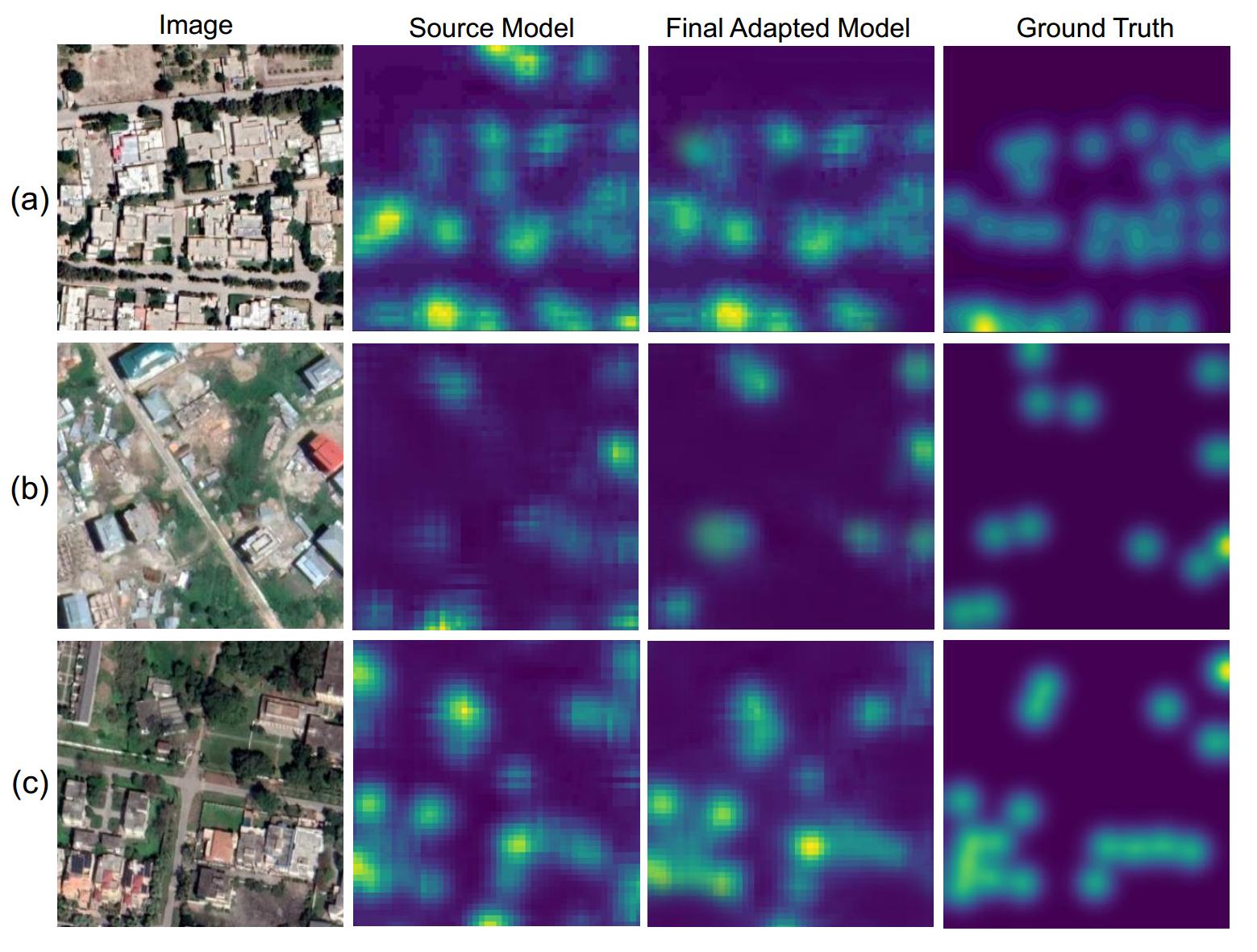}
\caption{From Left to Right: Input Image, distribution map predicted by source only trained model, distribution map predicted by the adapted model and Ground-truth. After adaptation, the predicted density map captures the localization information of the buildings much better than the ones produced by source only model.}
\label{fig: improvement}
\end{figure*} 

We evaluate our proposed approach on the testing set of IML-DAC  which consists of  988 images and on a subset of xView consisting of South Asian countries.\newline  
\noindent\textbf{Component-wise analysis of the proposed approach:} A detailed quantitative comparison of the proposed approach and its components is given in Table \ref{tab:results} and Table \ref{Tab:detailedMRE}. As indicated in Table \ref{tab:results}, the source-trained model suffers a significant decline in performance when tested on the target (IML-DAC) dataset. 
We detail both, MRE over the target and the reduction in MRE with respect to when the only source-trained model is used, as different loss functions are introduced. 
Where density map alignment results in a decrease in Mean Relative Error (MRE), the significant improvement comes with the introduction of \textit{Counting Consistency Constraints}, especially their combination. 
The final combination of all three losses results in the lowest counting error of $26.40\%$. A similar trend can also be seen while testing on south Asian regions of xView.
Note that DMA also contains MSE loss. 
\begin{table}[h!]
\centering
\caption{Comparison of mean relative error of all models on the target datasets. The first row is the model trained only on the source. Other rows represent adaptation by the indicated loss. Our final adapted model reduces error by approximately 7\% on the IML-DAC dataset and approximately 20\% on the South Asian subset of xView from the source trained model. MRE: lower is better. Reduction in Error: higher is better}
\label{tab:results}
\resizebox{.5\textwidth}{!}{%
\begin{tabular}{ccccc} 
\cline{2-5}
\multicolumn{1}{l}{} & \multicolumn{2}{c}{\begin{tabular}[c]{@{}c@{}}\textbf{Target: }\\\textbf{IML-DAC}\end{tabular}} & \multicolumn{2}{c}{\begin{tabular}[c]{@{}c@{}}\textbf{Target: }\\\textbf{xView (South Asian)}\end{tabular}} \\ 
\hline
\textbf{Experiments } & \textbf{\textbf{MRE}} & \begin{tabular}[c]{@{}c@{}}\textbf{\textbf{Reduction }}\\\textbf{\textbf{in Error}}\end{tabular} & \textbf{MRE} & \begin{tabular}[c]{@{}c@{}}\textbf{Reduction }\\\textbf{in Error}\end{tabular} \\ 
\hline
Source Trained Model & 33.14\% & - & 48.24\% & - \\ 
\hline
$\mathcal{L}_{DMA}$ & 31.97\% & 1.17\% & 37.40\% & 10.84\% \\ 
\hline
$\mathcal{L}_{CWI}$ & 29.45\% & 3.69\% & 33.30\% & 14.94\% \\ 
\hline
$\mathcal{L}_{CAI}$ & \textbf{\textbf{26.40\%}} & \textbf{\textbf{6.74\%}} & \textbf{28.41\%} & \textbf{19.83\%} \\
\hline
\end{tabular}
}
\end{table}

\begin{figure*}[t]
 \centering
\includegraphics[width=1.0\textwidth]{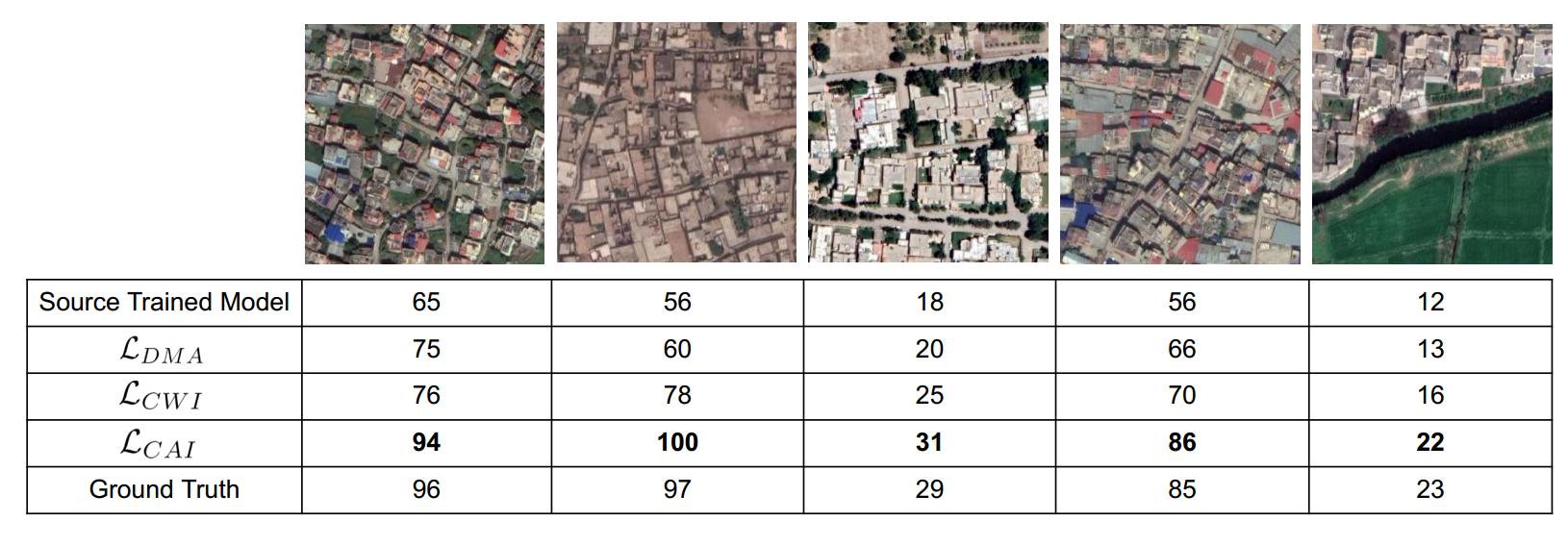}
\caption{Qualitative Results. Each column shows the improvement in building counting for the satellite image shown at the top of the column.}
\label{fig:count_improve}
\end{figure*}

To check the quality of density maps produced by our source and final adapted model, we have shown a detailed comparison in Figure \ref{fig: improvement}. It is noticeable that as the model is adapted from the source to the target dataset, the predicted density maps improve in quality, i.e, they are better able to locate and count the buildings. It is also worth noting that the density maps also become sharper as we adapt the model from the source to our target dataset.
Moreover, in Figure \ref{fig:count_improve} we observe the improvement in building counting for the satellite images using our proposed approaches.
\newline\newline

\begin{figure*}[ht]
\centering
\includegraphics[width=1.0\textwidth]{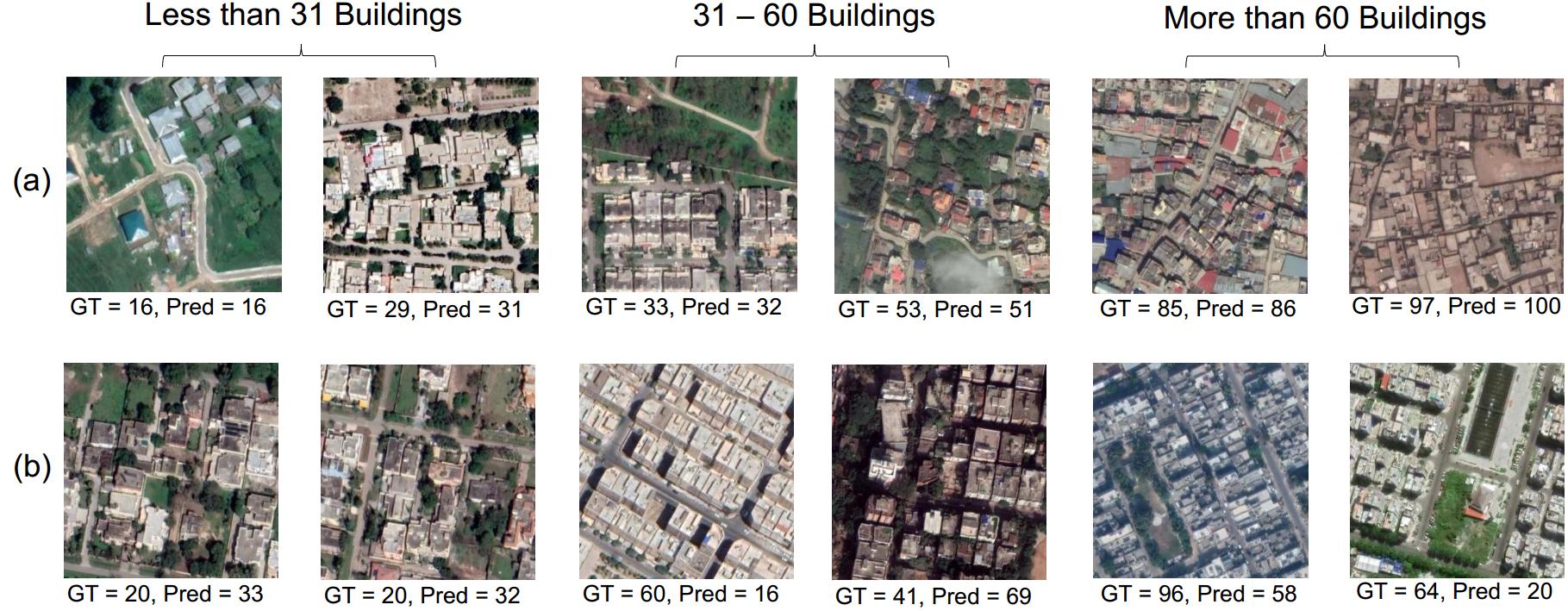}
\caption{Comparison of predicted and ground truth counts using our final adapted model on images containing a different range of buildings.  (a) depicts successfully predicted counts and (b) shows images where our model has failed to predict precise counts from images.}
\label{fig:adaptedcounts}
\end{figure*}

\noindent\textbf{Comparison with related works:} Since, to the best of our knowledge, we are the first one to address the problem of building counting across the regions,  {we have compared our methods with previous methods which have addressed domain adaptation in crowd counting. The method of \cite{hossain2020domain} minimized MMD loss between the source and target density maps, generated from crowded images, in a semi-supervised manner using three settings of the few-shot learning. In these three settings, 1, 5, and 10 labeled images respectively from the target domain were utilized while minimizing the MMD loss. We, however, have implemented their method in an unsupervised manner without using any labeled image of the target dataset. The work of \cite{li2019coda} also proposes to address domain adaptation in crowd counting by utilizing ranking and adversarial loss to adapt the target dataset to cater to different density distributions and various scales.

\begin{table}[ht]
\centering
\caption{In this table we compare our method with two of the existing works which deal with domain adaptation, but in crowd counting. Our final adapted model outperforms these models when tested on both the target datasets.}
\label{tab:results2}
\resizebox{.5\textwidth}{!}{%
\begin{tabular}{ccccc} 
\cline{2-5}
\multicolumn{1}{l}{} & \multicolumn{2}{c}{\begin{tabular}[c]{@{}c@{}}\textbf{Target: }\\\textbf{IML-DAC}\end{tabular}} & \multicolumn{2}{c}{\begin{tabular}[c]{@{}c@{}}\textbf{Target: }\\\textbf{xView (South Asian)}\end{tabular}} \\ 
\hline
\textbf{Experiments } & \textbf{\textbf{MRE}} & \begin{tabular}[c]{@{}c@{}}\textbf{\textbf{Reduction }}\\\textbf{\textbf{in Error}}\end{tabular} & \textbf{MRE} & \begin{tabular}[c]{@{}c@{}}\textbf{Reduction }\\\textbf{in Error}\end{tabular} \\ 
\hline
Source Trained Model & 33.14\% & - & 48.24\% & - \\ 
\hline
MMD \cite{hossain2020domain}  & 29.23\% & 3.91\% & 34.38\% & 13.86\% \\ 
\hline
CODA \cite{li2019coda} & 28.17\% & 4.97\% & 30.67\% & 17.57\% \\ 
\hline 
\textbf{Ours} & \textbf{\textbf{26.40\%}} & \textbf{\textbf{6.74\%}} & \textbf{28.41\%} & \textbf{19.83\%} \\
\hline
\end{tabular}
}
\end{table}

As demonstrated in Table \ref{tab:results2}, our proposed methodology outperforms both of these methods in terms of a higher reduction in error while testing on an unseen target domain.} \newline
\begin{table}[ht]
\centering
\caption{Comparison of mean relative error of all models across different ranges of buildings of target dataset (IML-DAC).}
\label{Tab:detailedMRE}
\resizebox{.5\textwidth}{!}{%
\begin{tabular}{ccccc} 
\hline
\multirow{2}{*}{\begin{tabular}[c]{@{}c@{}}Building \\Ranges\end{tabular}} & \begin{tabular}[c]{@{}c@{}}Source \\Trained Model\end{tabular} & $\mathcal{L}_{DMA}$ & $\mathcal{L}_{CWI}$ & $\mathcal{L}_{CAI}$ \\ 
\cline{2-5}
 & \textbf{MRE} & \textbf{MRE} & \textbf{MRE} & \textbf{MRE} \\ 
\hline
Less than 31 & 30.65 \% & 29.15 \% & 26.57 \% & \textbf{23.36 \%} \\ 
\hline
31 - 60 & 51.03 \% & 52.83 \% & 51.13 \% & \textbf{50.07 \%} \\ 
\hline
More than 60 & 55.04 \% & 55.98 \% & 53.52 \% & \textbf{50.78 \%} \\
\hline
\end{tabular}
}
\end{table}

\noindent\textbf{Results for different building count ranges:} To analyze the model's behavior as the density of the buildings changes, we report results on different ranges in the number of buildings. 
For this purpose, we have segregated our testing data (IML-DAC) into three divisions: (i) images containing less than 31 buildings, (ii) images containing 31 to 60 buildings, (iii) images containing more than 60 buildings, and  {compared the results of different models in Table \ref{Tab:detailedMRE}. We can observe that the count is more erroneous as we move to images with a high building count.}   In Figure \ref{fig:adaptedcounts}, we present the qualitative results on some of the images from these three ranges of buildings. Figure \ref{fig:adaptedcounts}(a) shows some images where count prediction is accurate, whereas Figure \ref{fig:adaptedcounts}(b) shows cases where predicted counts are inconsistent from the ground truths. 

\subsection{Ablation}

\begin{table}[h]
\centering
\caption{Comparison of M.R.E of source only model when trained on source dataset and tested on target dataset, with and without histogram equalization being part of preprocessing step. With histogram equalization generalization of the model improves.}
\label{tab:hist_result}
\resizebox{.5\textwidth}{!}{%
\begin{tabular}{ccc} 
\cline{2-3}
\multicolumn{1}{l}{}                                                            & \begin{tabular}[c]{@{}c@{}}\textbf{With }\\\textbf{Histogram Equalization}\end{tabular} & \begin{tabular}[c]{@{}c@{}}\textbf{\textbf{Without }}\\\textbf{\textbf{Histogram Equalization}}\end{tabular}  \\ 
\hline
\textbf{Experiments }                                                           & \textbf{\textbf{MRE}}                                                                   & \textbf{MRE}                                                                                                  \\ 
\hline

\begin{tabular}[c]{@{}c@{}}Source Trained Model\\Tested on IML-DAC\end{tabular} & \textbf{33.14}\%                                                                                 & 42.98\%                                                                                                       \\ 
\hline

\end{tabular}
}
\end{table}

\begin{figure*}[t]
\centering
\includegraphics[width=1.0\textwidth]{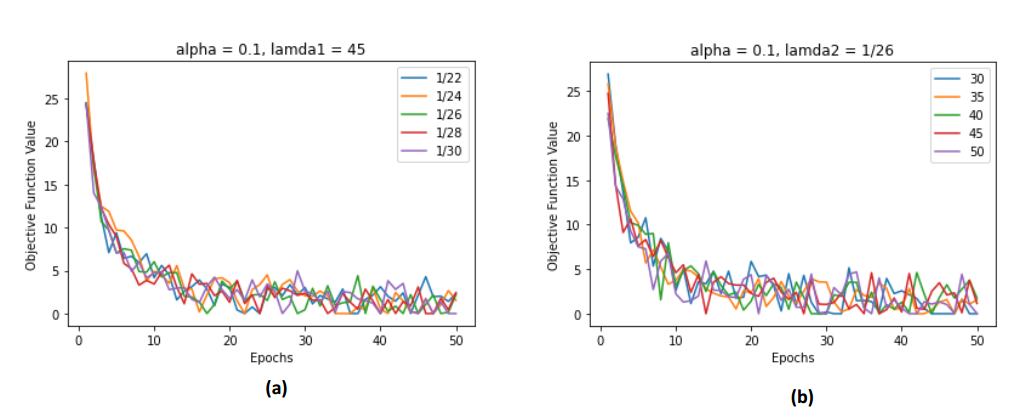}
\caption{{Convergence of performance during adaptation (a) $\lambda_1$ is fixed to 45 and $\lambda_2$ is being varied. (b) $\lambda_2$ is fixed to 1/26 and $\lambda_1$  is being varied.}} 
\label{fig:converge}
\end{figure*}
Histogram equalization is utilized to improve the contrast level of the images and is applied to each image separately. Figure \ref{fig:histcomparision}, qualitatively show the effect of this step.  
This preprocessing step improves the source trained model (that has not seen the target data) because it does not have to overcome the difference in contrast level in the two domains.
To show its effectiveness, we compared when the source-only trained model was trained on the source dataset and tested on the target dataset, without having this preprocessing step and when it is included. As shown in Table \ref{tab:results}, the histogram equalization source-only trained model performed 22.9\% better on the target dataset (IML-DAC).

{The Figure \ref{fig:converge} illustrates the convergence of our objective function. In these adaptation experiments shown in (a), we fixed alpha at 0.1, lambda1 ($\lambda_1$) at 45, and varied lambda2 ($\lambda_2$) from 1/22 to 1/30.
The figure in (b) shows alpha fixed at 0.1, lambda2 ($\lambda_2$) at 1/26 while lambda1 ($\lambda_1$) is varied from 30 to 55.
The given plots show the effect of these hyperparameters on performance convergence. \textit{For all these values we see convergence, indicating that hyperparameters are not too sensitive in these ranges.}}

\section{Limitations \& Future Directions} 

\label{sec:future}
 {
The current experiments on the target dataset were restricted to regions from South Asian countries only. The images covered not all but a few cities of these countries. For the future, a much larger dataset needs to be tagged and presented as standard for such studies, with special consideration to make it diverse and inclusive. 
In future work, we intend to include multi-task learning that exploits information such as the presence of roads or parks to improve the domain alignment and explainability component. 
}

\section{Conclusion} \label{sec:conclusion}
In this paper, we have addressed the challenging problem of cross-region building counting. We propose two counting consistency constraints to help direct the domain adaptation for the counting problem over the unlabeled target dataset. 
Exploiting the structure that should be there in the density map, we use adversarial learning to align the features across domains.
Furthermore, we have introduced a large-scale dataset based on satellite imagery consisting of regions belonging to various South Asian regions to validate our domain adaptation methodology. The quantitative results prove that adapting the source trained model using our approach of count consistency and output space adaptation can predict counts from the target dataset quite accurately. Our proposed approach acts as a benchmark in this setting as it does not require any labeled images from the target dataset, making the whole process of building counting computationally efficient and labor-saving. We reported an improvement of  {approximately 7\% on the IML-DAC dataset, and approximately 20\% on the South Asian subset of xView}  over the model trained on the source dataset only. The huge improvement South Asian subset of xView could be due to the reason that both developed and developing regions in xView are captured at the same resolution and with the same sensors. On the other hand, comparatively less improvement on the IML-DAC dataset could be due to the large domain shift and demonstrate that our dataset is more challenging. \newline\newline
\noindent\textbf{Conflict of Interest:} We wish to confirm that there are no known conflicts of interest associated with
this publication and there has been no significant financial support for this work that
could have influenced its outcome. We confirm that the manuscript has been read and
approved by all named authors and that there are no other persons who satisfied the
criteria for authorship but are not listed. We further confirm that the order of authors
listed in the manuscript has been approved by all of us. We confirm that we have given
due consideration to the protection of intellectual property associated with this work
and that there are no impediments to publication, including the timing of publication,
with respect to intellectual property. We confirm that we have followed the
regulations of our institutions concerning intellectual property. We understand that the corresponding author is the sole contact for the Editorial process (including the Editorial
Manager and direct communications with the office). He/she is responsible for communicating
with the other authors about progress, submissions of revisions, and final
approval of proofs. We confirm that we have provided a current, correct email address
which is accessible by the corresponding author. \newline

\noindent\textbf{Data Availability:} The target dataset IML-DAC that supports the findings of our methodology can be accessed at:  \href{https://github.com/intelligentMachines-ITU/domain-Adaptive-Building-Counting}{https://github.com / intelligentMachines-ITU/domain-Adaptive-Building-Counting}.

\bibliographystyle{plain}
\bibliography{sn-bibliography}

\end{document}